# Optimized Participation of Multiple Fusion Functions in Consensus Creation: An Evolutionary Approach

Elaheh Rashedi
Dept. of Electrical and Computer engineering
Isfahan University of Technology
Isfahan, Iran
elahehrashedi@gmail.com

Abdolreza Mirzaei
Dept. of Electrical and Computer engineering
Isfahan University of Technology
Isfahan, Iran
mirzaei@cc.iut.ac.ir

*Abstract* — **Recent studies show that ensemble methods enhance the stability and robustness of unsupervised learning. These approaches are successfully utilized to construct multiple clustering and combine them into a one representative consensus clustering of an improved quality. The quality of the consensus clustering is directly depended on fusion functions used in combination. In this article, the hierarchical clustering ensemble techniques are extended by introducing a new *evolutionary fusion function*. In the proposed method, multiple hierarchical clustering methods are generated via bagging. Thereafter, the consensus clustering is obtained using the search capability of genetic algorithm among different aggregated clustering methods made by different fusion functions. Putting some popular data sets to empirical study, the quality of the proposed method is compared with regular clustering ensembles. Experimental results demonstrate the accuracy improvement of the aggregated clustering results.**

## I. Introduction

Clustering is the problem of grouping data objects into some clusters so that the quality of intracluster similarity improves. There are extensive bodies of works done on clustering methods including ensemble based techniques which are proved to perform better than any single clustering methods in exchange for more complicated computations [1-4].

Cluster ensemble method can be defined as a two part problem: *creating* a set of $n$ clustering methods and *combining* them into one representative clustering of improved quality which maximized total agreement with $n$ clusterings [5-10]. The combination problem in clustering ensembles is more difficult than classifier ensembles. In the classification case, it is forthright to measure a classifier performance with respect to a training point, while in the clustering case there is a lack of knowledge about the label of the cluster to which a training point actually belongs [11].

Recent studies show that in many cases the quality of the clustering results directly depends on aggregation functions used in combination. Many aggregation techniques are introduced to combine ensembles, with a comprehensive body of works on partitional clusterings [12] and a few are introduced on hierarchical clusterings that are discussed in below.

*Partitional clustering fusing approaches*: Many consensus functions in these approaches are introduced for p-clustering ensembles which used a variety of mathematical tools [12]. Some are introduced as follows: information theory [13], fuzzy clustering [14], genetic algorithms [15-16], relabeling and voting [17-18], co-association matrix [19], graph and hypergraph [20], Mirkin distance [21], finite mixture models [22], locally adaptive clustering algorithm [23], kernel [24] and non-negative matrix factorization [25].

*Hierarchical clustering fusing approaches*: Some consensus functions in these approaches are introduced for hierarchical clustering ensembles. Among them, information theory based methods [26], fuzzy similarity relation based methods [27-28] and boosting based methods [29] can be found.

There is a potential of providing new consensus methods by mapping some partitioning based approaches to hierarchical ones. One of the mathematical tools which is not covered in hierarchical clustering area is the genetic algorithm. Having this motivation behind, this article reduces the general problem of clustering fusion to hierarchical clustering ones and focus on presenting a genetic based fusion function.

In this paper we proposed a new hierarchical clustering fusion function based on an evolutionary computing method that is genetic algorithm. In the mentioned approach, the consensus clustering is obtained using the search capability of genetic algorithm among different fusion functions. The main goal is performance improvement in term of accuracy using the best feature of each aggregated clustering made by each fusion function [12, 30].

The results of the proposed method are compared with the performance of the genetic based consensus algorithm with other fusion approaches for hierarchical clustering ensembles. Experimental results illustrate the effectiveness of the proposed method.

This paper is organized as follows. Section II presents a critical review on related works. In section III we formalize our idea and represent our combination scheme for constructing and aggregating ensemble results. Section IV demonstrates the implementations with some explanations on them. Finally, the conclusion results are given in section V.

## II. Preliminaries

### A. Matrix presentation of a hierarchical clustering

The arrangement of the clusters produced by hierarchical clustering is frequently presented by a tree diagram called dendrogram [31]. Dendrogram is usually mapped into a distance description matrix in which the $i^{th}$ row and $j^{th}$ column describe the distance between the $i^{th}$ and the $j^{th}$ object in the hierarchy of clusters. Various distance description were introduced [32]. One of them is Cophenetic Difference (*CD*) metric. The CD metric defines the distance between two samples as the lowest level of the hierarchy where these pairs are joined together. The hierarchy and the cophenetic matrix are two different forms of the same thing [33].

Likewise, the distance description matrices having the ultra metric property can be mapped into a dendrogram too. A matrix is said to be ultra metric when satisfying the following inequality:

$$d_{ij} \leq max(d_{ik}, d_{kj}) \quad \text{for } k = 1..n \qquad (1)$$

In which $n$ is the number of objects and $d_{ij}$ stands for the distance between the $i^{th}$ and the $j^{th}$ object. Distance description matrices felicitate the fusion functions' computation [34].

The proposed method encounters to the problem of hierarchical clustering combination by using matrix presentation of the hierarchies.

### B. Matrix fusion functions

The problem of hierarchical clustering combination becomes more feasible by mapping dendrograms into description matrices.

Obtaining description matrices, a fusion function can be applied on matrices and generates a consensus matrix. The final dendrogram is then derived from this matrix [31].

It should be noted that all aggregated matrices can not be mapped into a dendrogram unless they are ultrametric. Considering this situation, some aggregation functions are designed so that they keep this property and some other aggregators need to recover the final consensus matrix and so it becames ultrametric [28].

In this paper, the fusion function is supposed to be implemented on matrices which describe dendrograms.

### C. Cluster analysis and evaluation via matrices

Cluster analysis is a method of presenting the similarities and dissimilarities between pair of objects. Four General steps were introduced for cluster analysis [11, 33]:

1) Obtaining the data matrix: data matrix contains the attributes' value of each object.
2) Computing the similitude matrix: similitude matrix contains the degree of similarity between each pair of objects, called similitude coefficient. similitude coefficient falls into two categories, dissimilarity and similarity [16]. Euclidean distance coefficient is belongs to the second category.
3) Executing the clustering method: the similitude matrix turn into a hierarchy using a clustering method.
4) Computing the cophenetic correlation coefficient, *CPCC*: *CPCC* measures how well the hierarchy and the similitude matrix are similar [11, 16, 28-29, 35]. The comparison of a hierarchy and a matrix is not achievable, so the best thing is to convert the hierarchy into the equivalent cophenetic matrix which contains cophenetic distances.

Consequently, the evaluation of clustering results of our genetic based fusion function is done by utilizing the *CPCC* measure.

## III. Genetic Based Hierarchical Clustering Fusion Scheme

In this paper we proposed a new genetic based hierarchical clustering fusion technique. The mentioned approach is generally discussed below in six steps:

1) A primitive ensemble of *L* hierarchical clusterings, $\{H_1, H_2, ..., H_L\}$, is created. The ensemble of hierarchies can be generated by different algorithms or different runs of the same algorithm.
2) Thereafter, *CD* distance descriptor matrices corresponding to each hierarchy, $\{D_1, D_2, ..., D_L\}$, are obtained.
3) The secondary ensemble of $\{\acute{D}_1, \acute{D}_2, ..., \acute{D}_L\}$ is then generated by increasingly sorting the elements of primitive matrices so that the first matrix contains *minimum* distances and the last one contains *maximum* ones.
4) A genetic search algorithm is applied on secondary matrices to find the best weight of each matrix for participating in the agreement. The output of the genetic algorithm is a sequence of weights, $\{w_1, w_2, ..., w_L\}$, where $w_i$ is related to $\acute{D}_i$ and $\sum_{i=1}^{L} w_i = 1$.
5) Finding the best set of weights, the consensus matrix, $D^*$, is then generated by computing weighted average of sorted distance matrices. $D^*$ is computed as *eq. (2)*.

$$D^* = \sum_{i=1}^{L} w_i \times \acute{D}_i \qquad (2)$$

6) The consensus matrix is converted to an ultrametric one from which the consensus dendrogram is obtained.

More details are in below.

### A. Contribution of Rényi divergance measure in combination

Through information theory based methods, the normalized distance descriptor matrices are supposed to be a probability distribution function, *PDF*. Accordingly, the consensus matrix is defined to be a *PDF* with the most similarity to *PDF*s correspondent to individual ensemble matrices. Varieties of computational tools were presented to calculate the most similar consensus *PDF* among them the Rényi divergence measure can be found. According to this measure, the consensus *PDF* matrix is calculated as *eq. (3)*:

$$p_{ij}^* = \frac{1}{r}(\sum_{l=1}^{L}(p_{ij}^l)^{1-\alpha})^{\frac{1}{1-\alpha}} \quad (3)$$

In which $p_{ij}^*$ stand for the consensus *PDF* matrix, $p^*$, at the $i^{th}$ row and the $j^{th}$ column. Similarly, $p_{ij}^l$ is the $l^{th}$ individual matrix of the ensemble, $p^l$, at the $i^{th}$ row and the $j^{th}$ column. Finally, the *r* value is normalization constant. By setting α to some specific values, *eq. (1)* is changed into primitive functions; some of them are shown in table I.

TABLE I. DIFFERENT PRIMITIVE FUSION FUNCTIONS RELATED TO RÉNYI VIA DIFFERENT $\alpha$ VALUES

| $\alpha$ | Function | $p_{ij}^*$ |
|---|---|---|
| -∞ | Maximum | $\max_l p_{ij}^l$ |
| -1 | Euclidian length | $\frac{1}{r}\sqrt{\sum_{l=1}^{L}(p_{ij}^l)^2}$ |
| 0 | Arithmetic mean | $\frac{1}{r}\sum_{l=1}^{L} p_{ij}^l$ |
| 1 | Geometric mean | $\frac{1}{r}\prod_{l=1}^{L} p_{ij}^l$ |
| 2 | Harmonic mean | $\frac{1}{r}(\sum_{l=1}^{L} 1/p_{ij}^l)^{-1}$ |
| +∞ | Minimum | $\min_l p_{ij}^l$ |

It can be concluded from table I that when Rényi is applied on $\{Ď_1, Ď_2, .., Ď_L\}$, different outputs are supposed due to different values of $\alpha$. For example, if $\alpha = +\infty$, the consensus matrix will be $Ď_1$ as it contains the minimum distances, and similarly, if $\alpha = -\infty$, the consensus matrix will be $Ď_L$ as it contains the maximum distances.

### B. Contribution of Genetic Algorithm in combination

The proposed method uses the search capability of genetic algorithm to obtain the consensus clustering.

Generally, the population consists of *K* chromosomes, $P(t) = \{p_1, p_2, ..., p_K\}$. Each chromosome $p_k$ is a set of *L* weights, $\{w_1^k, w_2^k, ..., w_L^k\}$, which are associated with each matrix in the secondary ensemble $\{Ď_1, Ď_2, .., Ď_L\}$. For each $p_k$ we have $\sum_{i=1}^{L} w_i^k = 1$. Initially, the chromosomes are randomly generated using values between 0 to 1.

Based on *eq. (2)* which is formulated for obtaining the consensus clustering, the fitness associated to each chromosome $p_k$ is calculated as *eq. (4)* to *eq. (5)* in which $D^k$ is the consensus matrix determined by the chromosome $p_k$, and *E* is the Euclidian distance coefficient dissimilarity matrix of the original data.

$$D^k = \sum_{i=1}^{L} w_i^k \times Ď_i \quad (4)$$

$$fitness(p_k) = \left| \frac{\sum_{i<j}((D_{i,j}^k - \overline{D^k})(E_{i,j} - \bar{E}))}{\sqrt{\sum_{i<j}(D_{i,j}^k - \overline{D^k})^2 \sum_{i<j}(E_{i,j} - \bar{E})^2}} \right| \quad (5)$$

In *eq. (5)*, $D_{i,j}^k$ stands for the *CD* distance between the $i^{th}$ and the $j^{th}$ points in $D^k$, and $\overline{D^k}$ is the average of the $D_{i,j}^k$. similarly, $E_{i,j}$ stands for the ordinary Euclidean distance between the the $i^{th}$ and the $j^{th}$ points in the original data and $\bar{E}$ is the average of the $E_{i,j}$.

The fitness function is formulated so that it calculates the cophenetic correlation coefficient between the consensus matrix made by $p_k$ and the Euclidean distance matrix of the original data. As a result, the fitness determines which $p_k$ leads to a consensus clustering that is closer to what we are searching for. The higher is the value of fitness, the closer is the $p_k$ to the final sets of desired weights. So, the goal of the genetic algorithm is to search for the appropriate set of weights so that the clustering fitness is maximized.

Going after general genetic algorithms, crossover and mutation steps are also applied to obtain new population and the process continued until any termination criterion is achieved. Afterwards, the $p_k$ which leads to the highest fitness value is selected as the desired set of weights from which the target consensus matrix $D^*$ is generated.

### C. The overal Genetic based fusing scheme

The overall pseudo-code of the approach is in fig. 1.

In the following section, IV, the experimental results of applying the proposed method on real datasets are shown and discussed.

### IV. EXPERIMENTAL RESULTS AND DISCUSSIONS

Some experiments were conducted on real datasets in order to evaluate the performance of the combined hierarchical clustering result via the genetic based fusion function. Experimental setups, data sets and comparisons are discussed in the followings.

```
The Genetic based fusing scheme:

Input: distance descriptor matrices of the ensemble,
{D_1, D_2, ..., D_L}.
Output: the consensus matrix, D*.

1. Generate the secondary ensemble, {Ď_1, Ď_2, .., Ď_L} by
   increasingly sorting the elements of input matrices.
2. Begin the Genetic steps.
3. t = 1
4. Initialize the population P(1).
5. Evaluate the fitness of each individual in P(1),
   {p_1, p_2, ..., p_K}.
6. t = t + 1
7. Repeat on this generation until t < 100
   7.1. Select P(t) from P(t-1)
   7.2. Crossover P(t)
   7.3. Mutate P(t)
   7.4. Go to step 6.
8. Select p_k which satisfies max(fitness(p_k)).
9. Compute the consensus matrix D* using p_k =
   {w_1^k, w_2^k, ..., w_L^k}:
   D* = Σ_{i=1}^L w_i^k × Ď_i
10. Output D*
```

Figure 1. Pseudo code of the genetic based fusion function

### A. Experimental set up

In order to perform the application, some configurations should be set up.

First of all, a primitive ensemble of $L = 10$ hierarchical clustering methods, $\{H_1, H_2, ..., H_{10}\}$, is generated by different runs of the same hierarchical clustering algorithm namely *Single Linkage*. In order to make different hierarchies on the same data, the bagging idea is put to practical use [11]. Via bagging, a subsample of data points is participated in clustering instead of all. Here, we have chosen 80% of data points in each run of single linkage algorithm and created an ensemble of 10 hierarchies.

Thereafter, distance descriptor matrices corresponding to each hierarchy, $\{D_1, D_2, ..., D_L\}$, are obtained by utilizing Cophenetic Difference (*CD*) metric. The secondary sorted ensemble, $\{\acute{D}_1, \acute{D}_2, .., \acute{D}_L\}$ is then performed as declares in section III.

Afterward, the genetic algorithm is performed on the ensemble. So, the population size is set to $K = 100$, and also the mutation and the crossover rate are orderly set to 0.1 and 0.8.

Finally, by applying the genetic algorithm, the preferred set of weights are achieved and the target hierarchical clustering in then produced using *eq. (2)*. It should be noted that successive runs of presented algorithm, may produce different results, due to its heuristic nature. So the algorithm put to run for 10 times and the average result is documented here.

### B. Data sets

In this experiment eight datasets have been used from two popular databases, Real medical data sets [36] and UCI Repository of machine learning databases [37]. Datasets are chosen so that the numbers of data points are varied from $10^2$ to $10^4$ numbers. The Characteristics of datasets are shown in table II in more details.

TABLE II. CHARACTRISTICS OF DATASETS USED IN THIS EXPERIMENT

| Data set | #points | #features | reference |
|---|---|---|---|
| contraction | 98 | 27 | [36] |
| Wine | 178 | 13 | [37] |
| Wpbc | 198 | 32 | [37] |
| Weaning | 302 | 17 | [36] |
| Laryngeal2 | 692 | 16 | [36] |
| Vehicle | 846 | 18 | [37] |
| German | 1000 | 20 | [37] |
| Page_block | 5743 | 10 | [37] |

### C. Comparision with different fusion functions

In this part we compared the results of applying the presented genetic based algorithm with six different Rényi fusion functions, *maximum*, *Euclidian length*, *Average*, *Product*, *harmonic mean* and *minimum*. As it was previously mentioned, these fusion functions are conducted by setting $\alpha = \{-\infty, -1, \pm 0, +1, +2, +\infty\}$.

The Rényi functions and the genetic based fusion function are applied on primitive ensembles $\{D_1, D_2, ..., D_L\}$, and the combined hierarchies are generated. The comparison between performances of the final results is shown in term of *CPCC* metric in table III. The last row belongs to the proposed method. The maximum *CPCC* results are shown in bold face type.

### D. Statistical analysis

In this part, the best results on each dataset are determined by making a statistical analysis that is *t-test*.

The *t-test* analysis searches for the best quality method, *M*, and the other methods are compared to this one. In the next step of the *t-test*, methods which are not significantly different from *M* are determined. Thereafter, the winning frequency of each method is calculated as the percentage of the times which the mentioned method was determined.

In this experiment, *M* is the method which corresponds to maximum *CPCC* on each dataset and the significant level of comparison is set to 0.01.

Applying the *t-test*, the winning frequencies of all experimental methods are determined. Results are illustrated in fig. 2. The winning frequency of the presented approach is 75 percent across all experiments.

TABLE III. PERFORMANCE COMPARISON BETWEEN SIX RÉNYI BASED APPROACHES, (MAXIMUM, EUCLIDEAN LENGTH, ARITHMETIC MEAN, GEOMETRIC MEAN, HARMONIC MEAN AND MINIMUM), AND THE GENETIC BASED FUSING APPROACH. RESULTS ARE SHOWN IN TERM OF *CPCC* OF THE FINAL CONSENSUS HIERARCHICAL CLUSTERING.

|  |  | datasets | | | | | | | |
|---|---|---|---|---|---|---|---|---|---|
|  |  | contraction | Wine | Wpbc | Weaning | Laryngeal2 | Vehicle | German | Page_block |
| fusing methods | Maximum | 0.4787 | 0.4770 | **0.7434** | 0.6687 | **0.8737** | 0.5150 | 0.7393 | 0.9317 |
|  | Euclidian length | 0.4086 | 0.2604 | 0.6556 | 0.3152 | 0.7875 | 0.5028 | 0.6371 | 0.9238 |
|  | Arithmetic mean | 0.4707 | 0.5107 | **0.7460** | **0.6809** | 0.8474 | 0.5193 | 0.7873 | **0.9615** |
|  | Geometric mean | 0.3035 | 0.3286 | 0.4334 | 0.4828 | 0.3568 | 0.4836 | 0.5592 | 0.7582 |
|  | Harmonic mean | 0.4297 | 0.4616 | 0.7177 | 0.6553 | 0.8278 | 0.4986 | 0.7515 | 0.9405 |
|  | Minimum | 0.4791 | 0.5544 | 0.7387 | 0.6778 | 0.8421 | 0.5359 | **0.8022** | 0.9450 |
|  | Genetic Based Fusing | **0.7602** | **0.8805** | 0.7422 | 0.6838 | 0.8536 | **0.6919** | 0.7870 | **0.9623** |

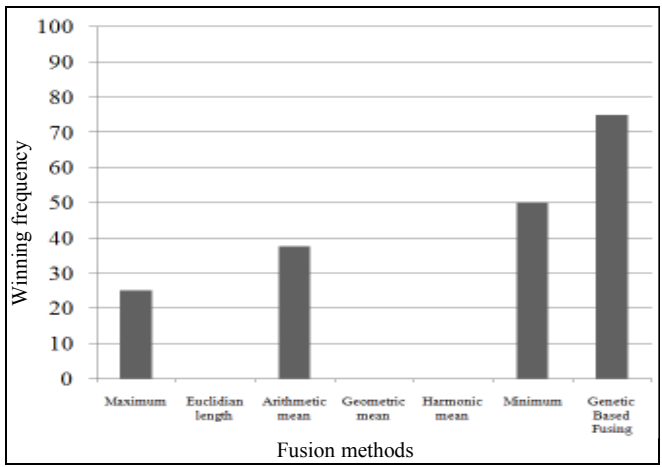

Figure 2. Winning frequencies of different fusion functions used in the experiments

According to fig. 2, it can be concluded that the genetic approach makes clustering of better quality in comparison with other primitive fusion functions in 75 percent of all cases.

## V. CONCLUSION

In this paper, a genetic based sheme for hierarchical clustering combination have been proposed. In this method, the problem of obtaining consensus hierarchical clustering is changed to an optimization problem which aims to participate multiple fusion functions in clustering aggregation. The projected problem is solved by using genetic algorithm which improves the accuracy by trying to use the best feature of different aggregated clusterings made by different fusion functions. Several experimental results indicate performance improvement of the consensus hierarchical clustering in terms of accuracy.

## VI. FURTHER WORKS

There is a potential for providing new consensus methods by mapping some partitioning based approaches to hierarchical ones. To the best of our knowledge, several mathematical tools are not covered in hierarchical clustering area, among them co-association matrix based methods and graph and hypergraph based methods can be found. These methods utilize graphs and matrices as their computational tools to create ensembles of partitions. In accordance to the close relation between graphs and hierarchical clusterings, further works may lead to presenting new ensemble approaches.